%% file: mono3d.tex
\documentclass[10pt,twocolumn,letterpaper]{article}

\usepackage{iccv}
\usepackage{times}
\usepackage{epsfig}
\usepackage{graphicx}
\usepackage{amsmath}
\usepackage{amssymb}
\usepackage{phonetic}
\usepackage{cite}

\usepackage{enumitem}
\usepackage[percent]{overpic}
\usepackage{multirow}
\usepackage{xcolor}
\usepackage[linesnumbered]{algorithm2e}

\usepackage[pagebackref=true,breaklinks=true,letterpaper=true,colorlinks,bookmarks=false]{hyperref}

 \iccvfinalcopy % *** Uncomment this line for the final submission

\def\iccvPaperID{6744} % *** Enter the ICCV Paper ID here

\newcommand\two{\text{2D}}
\newcommand\three{\text{3D}}

\definecolor{darkorange}{rgb}{0.773,0.353,.0667}
\definecolor{darkblue}{rgb}{0.184,0.322,.561}

\newcommand{\inv}{^{\raisebox{.2ex}{$\scriptscriptstyle-1$}}}

\newcommand{\Sim}{\mathord{\sim}}
\def\eqnvspace{{\vspace{-3mm}}}
\newcommand{\Paragraph}[1]{\vspace{1.25mm} \noindent \textbf{#1} \hspace{0mm}}

\newcommand{\SubSection}[1]{\vspace{-1.2mm} \subsection{#1} \vspace{-1.2mm}}

\newcommand{\xdownarrow}[1]{%
  {\left\downarrow\vbox to #1{}\right.\kern-\nulldelimiterspace}
}

\ificcvfinal\fi
\begin{document}

%%%%%%%%% TITLE
\title{ M3D-RPN: Monocular 3D Region Proposal Network for Object Detection }

\author{Garrick Brazil, Xiaoming Liu \\
Michigan State University, East Lansing MI \\
{\tt\small \{brazilga, liuxm\}@msu.edu}
}

\maketitle
\thispagestyle{empty}

%%%%%%%%% ABSTRACT
\begin{abstract}
Understanding the world in 3D is a critical component of urban autonomous driving. 
Generally, the combination of expensive LiDAR sensors and stereo RGB imaging has been paramount for successful 3D object detection algorithms, whereas monocular image-only methods experience drastically reduced performance. 
We propose to reduce the gap by reformulating the monocular 3D detection problem as a standalone 3D region proposal network.
We leverage the geometric relationship of 2D and 3D perspectives, allowing 3D boxes to utilize well-known and powerful convolutional features generated in the image-space. 
To help address the strenuous 3D parameter estimations, we further design depth-aware convolutional layers which enable location specific feature development and in consequence improved 3D scene understanding. 
Compared to prior work in monocular 3D detection, our method consists of only the proposed 3D region proposal network rather than relying on external networks, data, or multiple stages.
M3D-RPN is able to significantly improve the performance of both monocular 3D Object Detection and Bird's Eye View tasks within the KITTI urban autonomous driving dataset, while efficiently using a shared multi-class model.
\end{abstract}

\input{sec_1.tex}

\input{sec_2.tex}

\input{sec_3.tex}
\input{sec_4.tex}

\input{sec_5.tex}

{\small
\bibliographystyle{ieee_fullname}
\bibliography{egbib}
}

\end{document}

%% file: sec_1.tex
\section{Introduction}

Scene understanding in 3D plays a principal role in designing effective real-world systems such as in urban autonomous driving~\cite{Behl_2017_ICCV, chen2018lidar, Geiger2012CVPR} and robotics~\cite{guerry2017snapnet, tateno2017cnn}.
Currently, the foremost methods~\cite{chu2018surfconv, liang2018deep, qi2018frustum, shi2018pointrcnn, yang2018pixor} on 3D detection rely extensively on expensive LiDAR sensors to provide sparse depth data as input.
In comparison, monocular image-only 3D detection~\cite{chen2016monocular, chen20153d, mousavian20173d, xu2018multi} is considerably more difficult due to an inherent lack of depth cues.
As a consequence, the performance gap between LiDAR-based methods and monocular approaches remains substantial. 

Prior work on monocular 3D detection have each relied heavily on external state-of-the-art (SOTA) sub-networks, which are individually responsible for performing point cloud generation~\cite{chen20153d}, semantic segmentation~\cite{chen2016monocular}, 2D detection~\cite{mousavian20173d}, or depth estimation~\cite{xu2018multi}.
A downside to such approaches is an inherent disconnection in component learning as well as system complexity. 
Moreover, reliance on additional sub-networks can introduce persistent noise,  contributing to a limited upper-bound for the framework. 
%Moreover, the errors caused by sub-networks may escalate when propegated, thereby contributing to an overall limited upper-bound for the framework. 

\begin{figure}[t]
\vspace{-2mm}
\begin{center}
   \includegraphics[width=0.99\linewidth]{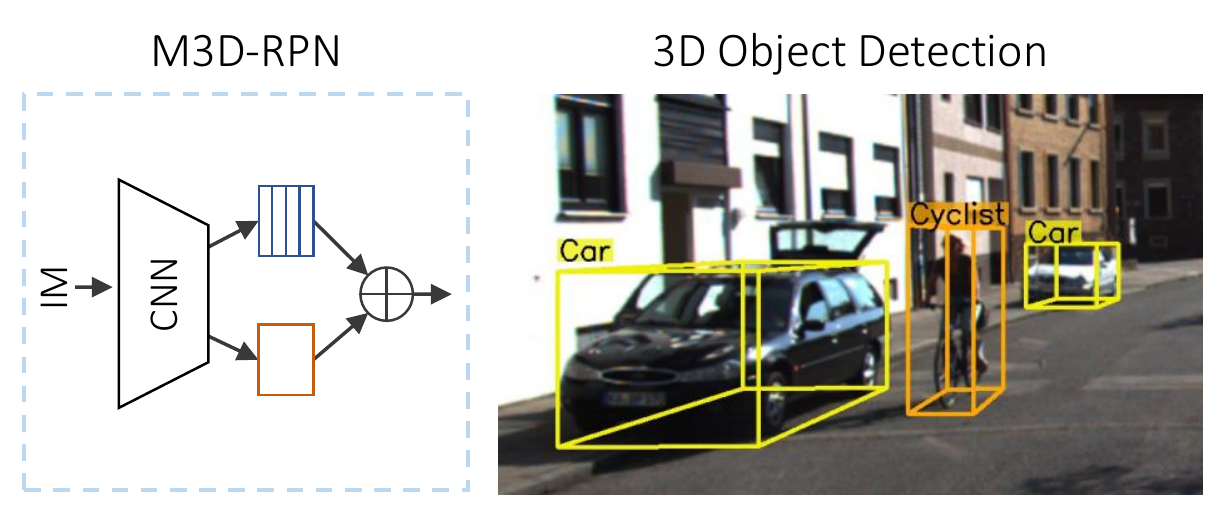}
\vspace{-1mm}
      \caption{
M3D-RPN uses a \textit{single} monocular 3D region proposal network with global convolution (\textcolor{darkorange}{orange}) and local depth-aware convolution (\textcolor{darkblue}{blue}) to predict multi-class 3D bounding boxes.
}
\label{fig:intro}
\end{center}\vspace{-6mm}
\end{figure}
In contrast, we propose a single end-to-end region proposal network for multi-class 3D object detection (Fig.~\ref{fig:intro}). 
%In effect, our network is completely standalone and does not require additional data or sub-networks, hence both performing more simply and efficiently. 
We observe that 2D object detection performs reasonably and continues to make rapid advances~\cite{cai2016unified, cai18cascadercnn, chen2018context, gao2018dynamic, kim2018parallel, ren2017accurate}. 
The 2D and 3D detection tasks each aim to ultimately classify all instances of an object; whereas they differ in the dimensionality of their localization targets.
Intuitively, we expect the power of 2D detection can be leveraged to guide and improve the performance of 3D detection, ideally within a unified framework rather than as separate components.
Hence, we propose to reformulate the 3D detection problem such that both 2D and 3D spaces utilize shared anchors and classification targets. 
In doing so, the 3D detector is naturally able to perform on par with the performance of its 2D counterpart, from the perspective of reliably classifying objects.
Therefore, the remaining challenge is reduced to 3D localization within the camera coordinate space. 

To address the remaining difficultly, we propose three key designs tailored to improve 3D estimation. 
Firstly, we formulate 3D anchors to function primarily within the image-space and initialize all anchors with prior statistics for each of its 3D parameters. 
Hence, each discretized anchor inherently has a strong prior for reasoning in 3D, based on the consistency of a fixed camera viewpoint and the correlation between 2D scale and 3D depth. 
%This formalization can be viewed as discretized classification in 2D space followed by  deviation for each 3D parameter. 
Secondly, we design a novel depth-aware convolutional layer which is able to learn spatially-aware features.
Traditionally, convolutional operations are preferred to be spatially-invariant~\cite{krizhevsky2012imagenet, lecun1999object} in order to detect objects at arbitrary image locations. 
However, while it is likely beneficial for low-level features, we show that high-level features improve when given increased awareness of their depth and while assuming a consistent camera scene geometry.
%Lastly, we integrate unsupervised elevation estimation which leverages the known perspective projection geometry to ray-trace depth using an adjusted ground plane. 
%The ray-traced depth is then weightedly fused with a direct anchor-based depth prediction, thereby enabling the network to conditionally use the scene geometry when necessary. 
Lastly, we optimize the orientation estimation $\theta$ using 3D~$\rightarrow$~2D projection consistency loss within a post-optimization algorithm. 
Hence, helping correct anomalies within $\theta$ estimation while assuming a reliable 2D bounding box. 

\noindent To summarize, our contributions are the following:
\begin{itemize}[noitemsep,topsep=1mm,label=$\bullet$]
\setlength\itemsep{1mm}
\item We formulate a standalone monocular 3D region proposal network (M3D-RPN) with a shared 2D and 3D detection space, while using prior statistics to serve as strong initialization for each 3D parameter.
\item We propose depth-aware convolution to improve the 3D parameter estimation, thereby enabling the network to learn more spatially-aware high-level features. 
\item We propose a simple orientation estimation post-optimization algorithm which uses 3D projections and 2D detections to improve the $\theta$ estimation.  
\item We achieve state-of-the-art performance on the urban KITTI~\cite{Geiger2012CVPR} benchmark for monocular Bird's Eye View and 3D Detection using a single multi-class network. %FIXME do you want to mention 2X-3X improvement?
\end{itemize}

%% file: sec_2.tex
\section{Related Work}
%Outline previous methods on 2D detection. 
%Note Faster R-CNN, YOLO, SSD for generic. 
%Note RRC, MS-CNN, etc for KITTI 
%Highlight the fact that most do not do ALL classes 
%with single model.
\Paragraph{2D Detection:}
Many works have addressed 2D detection in both generic~\cite{kong2018deep, li2018detnet, liu2016ssd, oksuz2018localization, redmon2016you} and urban scenes~\cite{gradient-feature-selection-for-online-boosting, cai2016unified, cai18cascadercnn, liu2018learning, ren2017accurate, yang2016exploit, zhou2018bi, brazil2017illuminating, pedestrian-detection-with-autoregressive-network-phases}.
Most recent frameworks are based on seminal work of Faster R-CNN~\cite{ren2015faster} due to the introduction of the region proposal network (RPN) as a highly effective method to efficiently generate object proposals.
The RPN functions as a sliding window detector to check for the existence of objects at every spatial location of an image which match with a set of predefined template shapes, referred to as anchors. 
Despite that the RPN was conceived to be a preliminary stage within Faster R-CNN, it is often demonstrated to have promising effectiveness being extended to a single-shot standalone detector~\cite{liu2016ssd, redmon2016you, womg2018tiny, zhang2018single}. 
Our framework builds upon the anchors of a RPN, specially designed to function in both the 2D and 3D spaces, and acting as a single-shot multi-class 3D detector. 
%However, state-of-the-art detectors in the urban domain generally adopt a one-class per model approach, as in ~\cite{?}. 
%In contrast, our method is equally performant when used to detect multi-class or single, demonstrating the robustness and better overall scaling capability.

\begin{figure}[t]
\vspace{-2mm}
\begin{center}
   \includegraphics[width=0.99\linewidth]{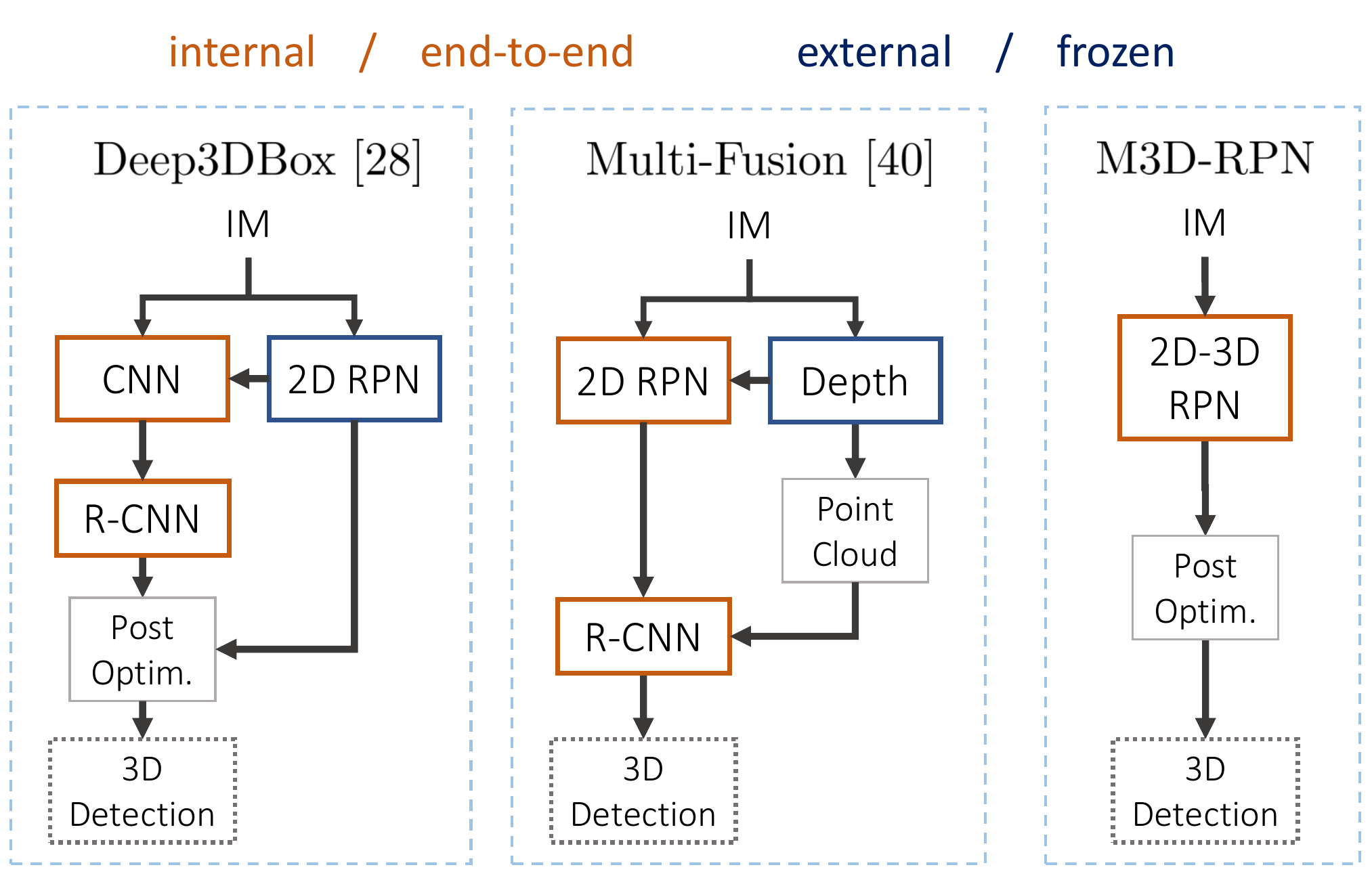}
\vspace{0mm}
      \caption{
Comparison of Deep3DBox~\cite{mousavian20173d} and Multi-Fusion~\cite{xu2018multi} with M3D-RPN. 
Notice that prior works are comprised of multiple internal stages (\textcolor{darkorange}{orange}), and external networks (\textcolor{darkblue}{blue}), whereas M3D-RPN is a {\it single-shot} network trained end-to-end.
}
\label{fig:prior_work}
\end{center}\vspace{-6mm}
\end{figure}

\Paragraph{LiDAR 3D Detection:}
The use of LiDAR data has proven to be essential input for SOTA frameworks~\cite{chen2017multi, chu2018surfconv, du2018general, liang2018deep, qi2018frustum, shi2018pointrcnn, yang2018pixor} for 3D object detection applied to urban scenes. 
Leading methods tend to process sparse point clouds from LiDAR points~\cite{qi2018frustum, shi2018pointrcnn, yang2018pixor} or project the point clouds into sets of 2D planes~\cite{chen2017multi, chu2018surfconv}.
While the LiDAR-based methods are generally high performing for a variety of 3D tasks, each is contingent on the availability of depth information generated from the LiDAR points or directly processed through point clouds. 
Hence, the methods are not applicable to camera-only applications as is the main purpose of our monocular 3D detection algorithm. 
%Each method is specifically tailored to efficiently process LiDAR point clouds and therefore cannot be implemented for on monocular RGB input alone. 
\begin{figure*}[t]
\vspace{-2mm}
\begin{center}
   \includegraphics[width=0.98\linewidth]{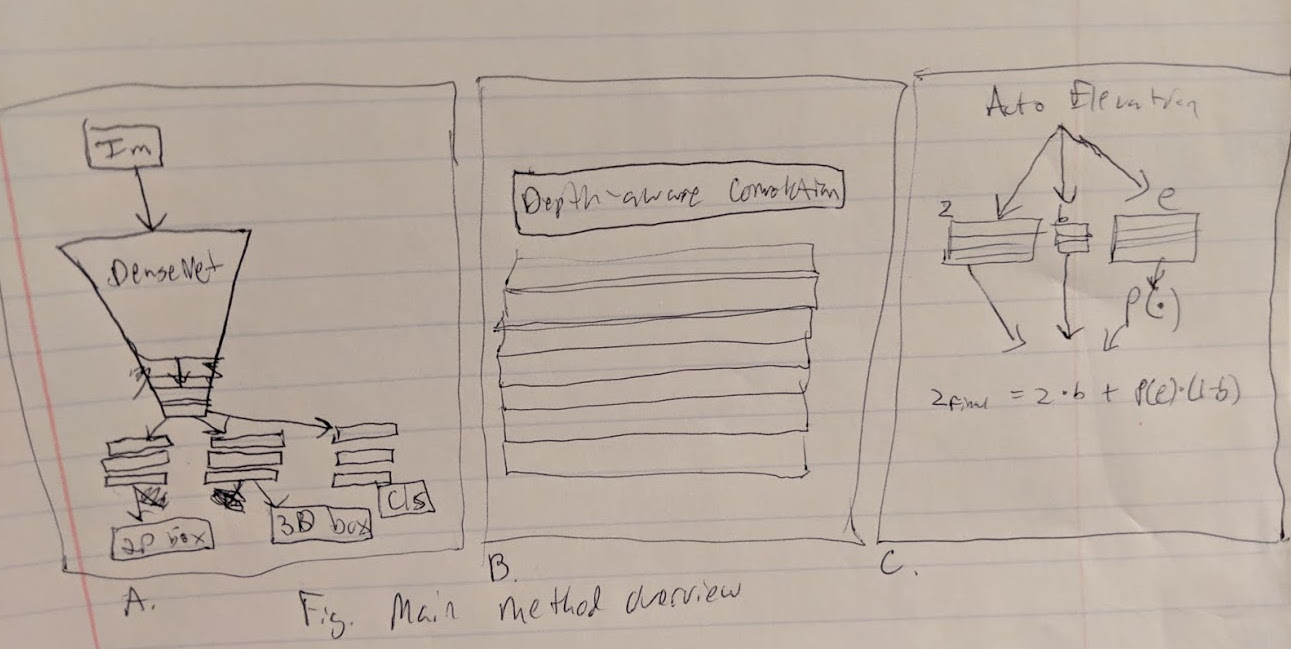}
\vspace{-2mm}
      \caption{
\textbf{Overview of M3D-RPN}. The proposed method consist of parallel paths for global (\textcolor{darkorange}{orange}) and local (\textcolor{darkblue}{blue}) feature extraction. 
The global features use regular spatial-invariant convolution, while the local features denote depth-aware convolution, as detailed right.
The depth-aware convolution uses non-shared kernels in the row-space $k_i$ for $i=1\dots b$, where $b$ denotes the total number of distinct bins.
To leverage both variants of features, we  weightedly combine each output parameter from the parallel paths. 
}

\label{fig:overview}
\end{center}\vspace{-6mm}
\end{figure*}

\Paragraph{Image-only 3D Detection:}
 3D detection using only image data is inherently challenging due to an overall lack of reliable depth information.
A common theme among SOTA image-based 3D detection methods~\cite{chen2016monocular, chen20153d, mousavian20173d, xu2018multi,monocular-video-based-trailer-coupler-detection-using-multiplexer-convolutional-neural-network} is to use a series of sub-networks to aid in detection.
%\cite{li2018stereo} propose to use 2D object detection and with stereo viewpoint classification in order to approximately estimate 3D boxes. 
%However, a limitation of this formulation is that it requires fixed 3D box dimension priors and relies on the the 2D center to be precisely consistent with the 3D projected counterpart. 
For instance,~\cite{chen20153d} uses a SOTA depth prediction with stereo processing to estimate point clouds.
Then 3D cuboids are exhaustively placed along the ground plane given a known camera projection matrix, and scored based upon the density of the cuboid region within the approximated point cloud. 
As a follow-up, \cite{chen2016monocular} adjusts the design from stereo to monocular by replacing the point cloud density heuristic with a combination of estimated semantic segmentation, instance segmentation, location, spatial context and shape priors, used while exhaustively classifying proposals on the ground plane. 

In recent work, \cite{mousavian20173d} uses an external SOTA object detector to generate 2D proposals then processes the cropped proposals within a deep neural network to estimate 3D dimensions and orientation.
Similar to our work, the relationship between 2D boxes and 3D boxes projected onto the image plane is then exploited in post-processing to solve for the 3D parameters.
However, our model directly predicts 3D parameters and thus only optimizes to improve $\theta$, which converges in $\Sim8$ iterations in practice compared with $64$ iterations in~\cite{mousavian20173d}. 
Xu \textit{et al.}~\cite{xu2018multi} utilize an additional network to predict a depth map which is subsequently used to estimate a LiDAR-like point cloud. 
The point clouds are then sampled using 2D bounding boxes generated from a separate 2D RPN. %, based on Faster R-CNN. 
Lastly, a R-CNN classifier receives an input vector consisting of the sampled point clouds and image features, to estimate the 3D box parameters. 

In contrast to prior work, we propose a \textit{single} network trained only with 3D boxes, as opposed to using a set of external networks, data sources, and composed of multiple stages. 
Each prior work \cite{chen2016monocular, chen20153d, mousavian20173d, xu2018multi} use external networks for at least one component of their framework, some of which have also been trained on external data.   
To the best of our knowledge, our method is the first to generate 2D and 3D object proposals simultaneously using a Monocular 3D Region Proposal Network (M3D-RPN).
In theory, M3D-RPN is complementary to prior work and may be used to replace the proposal generation stage. 
A comparison between our method and prior is further detailed in Fig.~\ref{fig:prior_work}.

%% file: sec_3.tex
\begin{figure*}[t]
\vspace{-2mm}
\begin{center}
   \includegraphics[width=0.99 \linewidth]{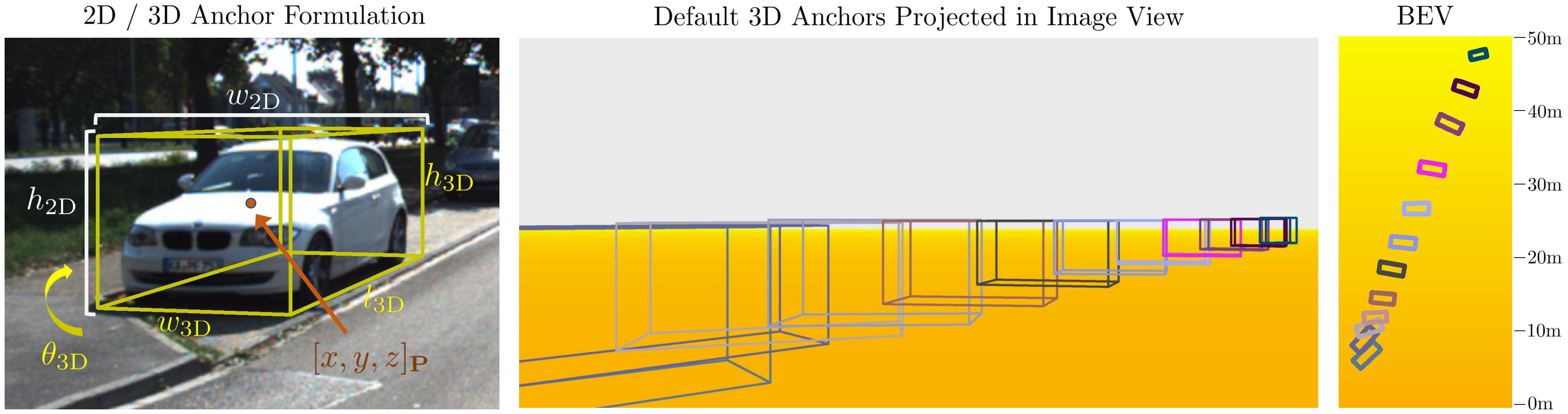}
\vspace{0mm}
      \caption{
\textbf{Anchor Formulation} and \textbf{Visualized 3D Anchors}. 
We depict each parameter of within the 2D~/~3D anchor formulation (left). 
We visualize the precomputed 3D priors when $12$ anchors are used after projection in the image view (middle) and Bird's Eye View (right). 
For visualization purposes only, we span anchors in specific $x_\three$ locations which best minimize overlap when viewed. 
}
\label{fig:anchors}
\end{center}\vspace{-5mm}
\end{figure*}

\section{M3D-RPN}
Our framework is comprised of three key components. %: anchor formulation, depth-aware convolution, and a 3D$\to$2D optimization. %, each critical to the performance of monocular 3D detection. 
First, we detail the overall formulation of our multi-class 3D region proposal network. %, which contributes the majority of the ability of our method. 
%We then outline the unsupervised elevation estimation which uses the ground-plane geometry to augment the $z$ depth estimation.
We then outline the details of depth-aware convolution and our collective network architecture.
Finally, we detail a simple, but effective, post-optimization algorithm for increased 3D$\to$2D consistency. 
We refer to our method as Monocular 3D Region Proposal Network (M3D-RPN), as illustrated in Fig.~\ref{fig:overview}.

\SubSection{Formulation}\label{sec:formulation}
The core foundation of our proposed framework is based upon the principles of the region proposal network (RPN) first proposed in Faster R-CNN~\cite{ren2015faster}, tailored for 3D. 
From a high-level, the region proposal network acts as sliding window detector which scans every spatial location of an input image for objects matching a set of predefined anchor templates. 
Then matches are regressed from the discretized anchors into continuous parameters of the estimated object.

\Paragraph{Anchor Definition:}
To simultaneously predict both the 2D and 3D boxes, each anchor template is defined using parameters of both spaces: $[w, h]_{\two}$, $z_\mathbf{P}$, and $[w, h, l, \theta]_{\three}$. 
For placing an anchor and defining the full 2D~/~3D box, a shared center pixel location $[x, y]_\mathbf{P}$ must be specified. 
The parameters denoted as 2D are used as provided in pixel coordinates.
We encode the depth parameter $z_\mathbf{P}$ by projecting the 3D center location $[x, y, z]_\three$ in camera coordinates into the image given a known projection matrix $\mathbf{P} \in \mathbb{R}^{3\times4}$ as 

\begin{equation}
\eqnvspace
\begin{bmatrix}
    x \cdot z \\
    y \cdot z \\
    z  \\
  \end{bmatrix}_\text{P} =
\mathbf{P} \cdot \begin{bmatrix}
    ~x~  \\
    ~y~  \\
    ~z~ \\
    ~1~ \\
  \end{bmatrix}_\three.
\eqnvspace\label{eqn:proj}
\end{equation}
The $\theta_\three$ represents the observation viewing angle~\cite{Geiger2012CVPR}.
Compared to the Y-axis rotation in the camera coordinate system, the observation angle accounts for the relative orientation of the object with respect to the camera viewing angle rather than the Bird's Eye View (BEV) of the ground plane. 
Therefore, the viewing angle is intuitively more meaningful to estimate when dealing with image features. 
We encode the remaining 3D dimensions $[w, h, l]_\three$ as given in the camera coordinate system.

The mean statistic for each $z_\mathbf{P}$ and $[w, h, l, \theta]_\three$ is pre-computed for each anchor individually, which acts as strong prior to ease the difficultly in estimating 3D parameters.
Specifically, for each anchor we use the statistics across all matching ground truths which have $\geq 0.5$ intersection over union (IoU) with the bounding box of the corresponding $[w, h]_\two$ anchor.
As a result, the anchors represent discretized templates where the 3D priors can be leveraged as a strong initial guess, thereby assuming a reasonably consistent scene geometry. 
We visualize the anchor formulation as well as precomputed 3D priors in Fig.~\ref{fig:anchors}.  

\Paragraph{3D Detection:}
Our model predicts output feature maps per anchor for
$c, [t_x, t_y, t_w, t_h]_\two, [t_x, t_y, t_z]_\mathbf{P}, [t_w, t_h, t_l, t_\theta]_\three$. 
Let us denote $n_a$ the number of anchors, $n_c$ the number of classes, and $h \times w$ the feature map resolution. 
As such, the total number of box outputs is denoted $n_b = w\times h \times n_a$, spanned at each pixel location $[x,y]_\mathbf{P} \in \mathbb{R}^{w \times h}$ per anchor.
The first output $c$ represents the shared classification prediction  of size $n_a \times n_c \times h \times w$, whereas each other output has size $n_a \times h \times w$. 
%We next detail the proposed formulation for inferring 3D detection from these parameters. 
%The first output ${c}$ denotes the shared classification prediction. 
The outputs of $[t_x, t_y, t_w, t_h]_\two$ represent the 2D bounding box transformation, which we collectively refer to as $b_\two$.
Following~\cite{ren2015faster}, the bounding box transformation is applied to an anchor with $[{w}, {h}]_\two$ as:
\begin{align}
\eqnvspace
\begin{split}
{x}'_\two = x_\mathbf{P} + t_{x_\two} \cdot {w}_\two,~~~~~~~~&{w}'_\two = \exp(t_{w_\two}) \cdot {w}_\two, \\
{y}'_\two = y_\mathbf{P} + ~t_{y_\two} \cdot {h}_\two,~~~~~~~~&{h}'_\two = \exp(t_{h_\two~}) \cdot {h}_\two, \\
\end{split}
\eqnvspace
\end{align}
where $x_\mathbf{P}$ and $y_\mathbf{P}$ denote spatial center location of each box.
The transformed box ${b}'_\two$ is thus defined as $[x, y, w, h]'_\two$.
The following $7$ outputs represent transformations denoting the projected center $[t_x, t_y, t_z]_\mathbf{P}$, dimensions $[t_w, t_h, t_l]_\three$ and orientation $t_{\theta_\three}$, which we collectively refer to as $b_\three$.
Similar to 2D, the transformation is applied to an anchor with parameters $[{w}, {h}]_\two$, $z_\mathbf{P}$, and $[{w}, {h}, {l}, {\theta}]_\three$ as follows:
\begin{align}
\eqnvspace
\begin{split}
x'_\mathbf{P} = x_\mathbf{P} + t_{x_\mathbf{P}} \cdot {w}_\two,~~~~~~~~~&w'_\three = \exp(t_{w_\three}) \cdot {w}_\three, \\
y'_\mathbf{P} = y_\mathbf{P} + ~t_{y_\mathbf{P}} \cdot {h}_\two,~~~~~~~~~&h'_\three = \exp(t_{h_\three})\cdot {h}_\three, \\
z'_\mathbf{P} = t_{z_\mathbf{P}} + ~{z}_\mathbf{P},~~~~~~~~~~~~~~~~&~l'_\three = \exp(t_{l_\three})\cdot {l}_\three, \\
\theta'_\three = t_{\theta_\three} + ~{\theta}_\three.~~~~~~~~~~~~~~~&~ \\
\end{split}
\eqnvspace \label{eqn:3d_transform}
\end{align}
Hence, $b'_\three$ is then denoted as $[x, y, z]'_\mathbf{P}$ and $[w, h, l, \theta]'_\three$.
As described, we estimate the projected 3D center rather than camera coordinates to better cope with the convolutional features based exclusively in the image space. 
Therefore, during inference we back-project the projected 3D center location from the image space $[x, y, z]'_\mathbf{P}$ to camera coordinates $[x, y, z]'_\three$ by using the inverse of Eqn.~\ref{eqn:proj}. % a known camera projection matrix $\bf{P}$ and the following equation:
%\begin{equation}
%\eqnvspace
%\mathbf{P}\inv \cdot \begin{bmatrix}
%    x \cdot z \\
%    y \cdot z \\
%    z  \\
%  \end{bmatrix}'_\text{P} =
%\begin{bmatrix}
%    x  \\
%    y  \\
%    z \\
%  \end{bmatrix}_\three.
%\eqnvspace
%\end{equation}

\Paragraph{Loss Definition:}
The network loss of our framework is formed as a multi-task learning problem composed of classification $L_{c}$ and a box regression loss for 2D and 3D, respectfully denoted as $L_{b_\two}$ and $L_{b_\three}$. %, and is computed for each generated 2D~/~3D box.
For each generated box, we check if there exists a ground truth with at least $\geq 0.5$ IoU, as in~\cite{ren2015faster}.
If yes then we use the best matched ground truth for each generated box to define a target with ${\tau}$ class index, 2D box $\hat{b}_\two$, and 3D box $\hat{b}_\three$.
Otherwise, $\tau$ is assigned to the catch-all background class and bounding box regression is ignored. 
A softmax-based multinomial logistic loss is used to supervise for $L_\text{c}$ defined as:
\begin{equation}
\eqnvspace
L_\text{c} = -\log\left(\frac{\exp(c_{{\tau}})}{\Sigma_i^{n_c} \exp(c_i)}\right). 
\eqnvspace
\end{equation}
%where $\hat{c}$ denotes the target $t_{cls}$ and $j$ indexes from $1\dots n_c$.
We use a negative logistic loss applied to the IoU between the matched ground truth box $\hat{b}_\two$ and the transformed $b'_{\two}$ for $L_{b_{\two}}$, similar to~\cite{wang2017repulsion, yu2016unitbox}, defined as:
\begin{equation}
\eqnvspace
L_{b_{\two}} = -\log\left(\text{IoU}(b'_{\two}, \hat{b}_\two)\right).
\eqnvspace
\end{equation}
The remaining 3D bounding box parameters are each optimized using a Smooth $L_1$~\cite{girshick2015fast} regression loss applied to the transformations $b_\three$ and the ground truth transformations $\hat{g}_\three$ (generated using $\hat{b}_\three$ following the inverse of Eqn.~\ref{eqn:3d_transform}):
\begin{equation}
\eqnvspace
L_{b_\three} = \text{Smooth}L_1(b_\three, \hat{g}_\three).
\eqnvspace
\end{equation}
Hence, the overall multi-task network loss $L$, including regularization weights $\lambda_1$ and $\lambda_2$, is denoted as:
\begin{equation}
\eqnvspace
L = L_{\text{c}} + \lambda_1 L_{b_\two} + \lambda_2 L_{b_\three}.
\eqnvspace
\end{equation}

\SubSection{Depth-aware Convolution}

Spatial-invariant convolution has been a principal operation for deep neural networks in computer vision~\cite{krizhevsky2012imagenet, lecun1999object}.  
We expect that low-level features in the early layers of a network can reasonably be shared and are otherwise invariant to depth or object scale. 
However, we intuitively expect that {high-level} features related to 3D scene understanding are dependent on depth when a fixed camera view is assumed.
As such, we propose depth-aware convolution as a means to improve the spatial-awareness of high-level features within the region proposal network, as illustrated in Fig.~\ref{fig:overview}. 

The depth-aware convolution layer can be loosely summarized as regular 2D convolution where a set of discretized depths are able to learn non-shared weights and features. 
We introduce a hyperparameter $b$ denoting the number of row-wise bins to separate a feature map into, where each learns a unique kernel $k$.
%The number of bins is inherently bounded by range $[1, h]$.
%The setting of $b = 1$ is the base equivalent of a normal convolution layer. 
%Moreover, the setting $b = h$ is the absolute maximum number of bins a $h\times w$ feature map can be grouped into.
%For each convolutional kernel $k$ in normal convolution, there are instead $k_i$ for $i=1 \dots b$ which are specifically applied to their respective bins. 
In effect, depth-aware kernels enable the network to develop location specific features and biases for each bin region, ideally to exploit the geometric consistency of a fixed viewpoint within urban scenes.
For instance, high-level semantic features, such as encoding a feature for a large wheel to detect a car, are valuable at close depths but not generally at far depths.
Similarly, we intuitively expect features related to 3D scene understanding are inherently related to their row-wise image position. % when using fixed cameras. % system.  %, just as features encoding far depth would not be useful for the later bins.  

An obvious drawback to using depth-aware convolution is the increase of memory footprint for a given layer by $\times b$.  
However, the total theoretical FLOPS to perform convolution remains consistent regardless of whether kernels are shared. 
We implement the depth-aware convolution layer in PyTorch~\cite{paszke2017automatic} by unfolding a layer $L$ into $b$ padded bins then re-purposing the group convolution operation to perform efficient parallel operations on a GPU\footnote{In practice, we observe a $10-20\%$ overhead for reshaping when implemented with parallel group convolution in PyTorch~\cite{paszke2017automatic}.}. 
\begin{table*}[t!]
\begin{center}
%\small
\setlength\tabcolsep{5.25pt}
  \resizebox{\textwidth}{!}{  
\begin{tabular}{l | c | c c c  | c c c}
\hline
\multirow{2}{*}{ } & \multirow{2}{*}{Type} & \multicolumn{3}{c|}{IoU $\geq 0.7$~~\small{[val1~/~val2~/~test]}} & \multicolumn{3}{c}{IoU $\geq 0.5$ \small{[val1~/~val2]}}  \\ 
& & Easy & Mod & Hard & Easy & Mod & Hard \\ 
\hline
Mono3D~\cite{chen2016monocular}  		& Mono 		& $~~5.22$ /~~~~~-~~~~~/~~~~~-~~~~	& $~~5.19$ /~~~~~-~~~~~/~~~~~-~~~~ 	& $~~4.13$ /~~~~~-~~~~~/~~~~~-~~~~ & $30.50$ /~~~~~-~~~~  & $22.39$ /~~~~~-~~~~ & $19.16$ /~~~~~-~~~~ \\ 
3DOP~\cite{chen20153d}  			& Stereo 	& $12.63$ /~~~~~-~~~~~/~~~~~-~~~~  	& $~~9.49$ /~~~~~-~~~~~/~~~~~-~~~~ 	& $~~7.59$ /~~~~~-~~~~~/~~~~~-~~~~ & $55.04$ /~~~~~-~~~~  & $41.25$ /~~~~~-~~~~ & $34.55$ /~~~~~-~~~~ \\  
Deep3DBox~\cite{mousavian20173d}  		& Mono 		& ~~~~-~~~~~/ $~~9.99$ /~~~~~-~~~~ 	&  ~~~~-~~~~~/ $~~7.71$ /~~~~~-~~~~	& ~~~~-~~~~~/ $~~5.30$ /~~~~~-~~~~ & ~~~~-~~~~~/ $30.02$  & ~~~~-~~~~~/ $23.77$ & ~~~~-~~~~~/ $18.83$ \\ 
Multi-Fusion~\cite{xu2018multi}  	& Mono 		& $22.03$ / $19.20$ / $13.73$	& $13.63$ / $12.17$ / $9.62$ & $11.60$ / $10.89$ / $8.22$	   & $55.02$ / $54.18$ 	  & $36.73$ / $38.06$ 	& $31.27$ / $31.46$ \\ 
\hline\hline
M3D-RPN  & Mono 		& $25.94$ / $26.86$ / $26.43$ 	& $21.18$ / $21.15$ / $18.36$ 	& $17.90$ / $17.14$ / $16.24$		& $55.37$ / $55.87$	  & $42.49$ / $41.36$ 	& $35.29$ / $34.08$ \\ 
\hline
\end{tabular}
}
\end{center}
\caption{
\textbf{Bird's Eye View}. Comparison of our method to image-only 3D localization frameworks on the Bird's Eye View task (AP$_\text{BEV}$). 
%We detail the AP$_\text{BEV}$ performance for the car class on $3$ dataset splits when computed using each difficult protocol.
}\label{tab:BEV}
\vspace{-2mm}
\end{table*}

\begin{table*}[t!]
\begin{center}
%\small
\setlength\tabcolsep{5.25pt}
  \resizebox{\textwidth}{!}{  
\begin{tabular}{l | c | c c c  | c c c}
\hline
\multirow{2}{*}{ } & \multirow{2}{*}{Type} & \multicolumn{3}{c|}{IoU $\geq 0.7$~~\small{[val1~/~val2~/~test]}} & \multicolumn{3}{c}{IoU $\geq 0.5$ \small{[val1~/~val2]}}  \\ 
& & Easy & Mod & Hard & Easy & Mod & Hard \\ 
\hline
Mono3D~\cite{chen2016monocular}  		& Mono 		& $~~~2.53$ /~~~~~-~~~~~/~~~~~-~~~~	& $~~2.31$ /~~~~~-~~~~~/~~~~~-~~~~ 	& $~~2.31$ /~~~~~-~~~~~/~~~~~-~~~~ & $25.19$ /~~~~~-~~~~	& $18.20$ /~~~~~-~~~~ 	& $15.52$ /~~~~~-~~~~ 	\\ 
3DOP~\cite{chen20153d}  			& Stereo 	& $~~~6.55$ /~~~~~-~~~~~/~~~~~-~~~~	& $~~5.07$ /~~~~~-~~~~~/~~~~~-~~~~ 	& $~~4.10$ /~~~~~-~~~~~/~~~~~-~~~~ & $46.04$ /~~~~~-~~~~ 	& $34.63$ /~~~~~-~~~~ 	& $30.09$ /~~~~~-~~~~  \\  
Deep3DBox~\cite{mousavian20173d}  		& Mono 		& ~~~~-~~~~~/$~~5.85$ /~~~~~-~~~~ 	&  ~~~~-~~~~~/ $~~4.10$ /~~~~~-~~~~	& ~~~~-~~~~~/ $~~3.84$ /~~~~~-~~~~ & ~~~~-~~~~~/ $27.04$ 	& ~~~~-~~~~~/ $20.55$ 	& ~~~~-~~~~~/ $15.88$ \\ 
Multi-Fusion~\cite{xu2018multi}  	& Mono 		& $10.53$ / $~~7.85$ / $7.08$ 	& $~~5.69$ / $~~5.39~$ / $5.18$	& $~~5.39$ / $~~4.73$ / $4.68$  & $47.88$ / $45.57$ 	& $29.48$ / $30.03$ 	& $26.44$ / $23.95$ \\ 
\hline\hline
M3D-RPN  & Mono 		& $20.27$ / $20.40$ / $20.65$ & $17.06$ / $16.48$ / $15.70$ & $15.21$ / $13.34$ / $13.32$  & $48.96$ / $49.89$ 	& $39.57$ / $36.14$ 	& $33.01$ / $28.98$ \\ 
\hline
\end{tabular}
}
\end{center}
\caption{
\textbf{3D Detection}. Comparison of our method to image-only 3D localization frameworks on the 3D Detection task (AP$_\text{3D}$). 
}\label{tab:3D}
\vspace{-2mm}
\end{table*}

%As such, the depth-aware convolution can act as a drop-in replacement for regular convolution where its inputs are the $b$, $n_\text{in}$, $n_\text{out}$, $k$, and $p$, where $n$ denotes the number of features incoming and outgoing and $p$ the padding size. %FIXME definition of k?
%Further, any convolutional layer with $b_0$ bins can be translated into a higher-group depth-aware layer with $b_1$, such that $b_1 \geq b_0$, by duplicating weights as initialization. 
%We utilize depth-aware convolutional as a drop-in replacement for each layer in the local path of our network (as shown in Fig.~\ref{fig:overview}).
%\Paragraph{Stacking depth-aware convolutional layers:} While we find it helpful to stack depth-aware layers to increase the power of spatially-variant features, it is worth acknowledging the substantial increase in complexity to converge or generalize. 
%In practice it may be necessary to pre-train with regular convolution then do weight-transfer into a depth-aware convolutional layer \textit{or} perform high levels of data augmentation. 

\begin{algorithm}[t]
\textbf{Input:} $b'_\two, [x, y, z]'_\mathbf{P}, [w, h, l, \theta]'_\three, \sigma, \beta, \gamma$   \smallskip \\
  $\rho \gets \text{box-project}([x, y, z]_\mathbf{P}, [w, h, l, \theta - \sigma]_\three)$ \\
  $\eta  \gets L_1(b'_\two,~\rho)$ \smallskip \\
 \While{$\sigma \geq \beta$}{  	
 \smallskip
	$\rho^{-} \gets \text{box-project}([x, y, z]_\mathbf{P}, [w, h, l, \theta - \sigma]_\three)$ \\
	$\rho^{+} \gets \text{box-project}([x, y, z]_\mathbf{P}, [w, h, l, \theta + \sigma]_\three)$ \smallskip \\
  	$loss^{-}  \gets L_1(b'_\two,~\rho^{-})$ \\
	$loss^{+}  \gets L_1(b'_\two,~\rho^{+})$\smallskip
	
%	\small{\tcp{reduce}}
	\uIf{$\min(loss^{-}, loss^{+}) > \eta $}{
		$\sigma \gets \sigma \cdot \gamma;$ \smallskip
	}
%	\small{\tcp{update $-$}}
	\uElseIf{$loss^{-} < loss^{+}$}{
    	$\theta \gets \theta - \sigma;$ \\
    	$ \eta \gets loss^{-}$ \smallskip
  	}
%	\small{\tcp{update $+$}}
\nl	\Else{
\nl    	$\theta \gets \theta + \sigma;$ \\
    	$ \eta \gets loss^{+}$
  	}
 }
 \caption{\textbf{Post 3D$\rightarrow$2D Algorithm.} The algorithm takes input of 2D~/~3D box $b'_\two, [x, y, z]'_\mathbf{P}, [w, h, l, \theta]'_\three$, step size $\sigma$, termination $\beta$, and decay $\gamma$ parameters, then iteratively tunes $\theta$ via $L_1$ corner consistency loss. }
 \label{hillclimb}
\end{algorithm}\setlength{\textfloatsep}{10pt}
\SubSection{Network Architecture}
The backbone of our network uses DenseNet-121~\cite{huang2017densely}.
We remove the final pooling layer to keep the network stride at $16$, then dilate each convolutional layer in the last DenseBlock by a factor of $2$ to obtain a greater field-of-view.
%, hence given an input resolution $W\times H$ then output is $\frac{W}{16}\times\frac{H}{16}$.
%Therefore, given an input image is of resolution $W\times H$, the corresponding output resolution is $w = \frac{W}{16}$, $h = \frac{H}{16}$. 

We connect two parallel paths at the end of the backbone network. 
The first path uses regular convolution where kernels are shared spatially, which we refer to as global. 
The second path exclusively uses depth-aware convolution and is referred to as local.
%The primary distinction between the paths is that global utilizes regular spatial-invariant convolution whereas the local path utilizes \textit{depth-aware convolution} to learn location specific features as detailed in Sec.\ref{?}.
For each path, we append a proposal feature extraction layer using its respective convolution operation to generate $\mathbf{F}_\text{global}$ and $\mathbf{F}_\text{local}$. 
Each feature extraction layer generates $512$ features using a $3\times3$ kernel with $1$ padding and is followed by a ReLU non-linear activation.
We then connect the $12$ outputs to each $\mathbf{F}$ corresponding to $c, [t_x, t_y, t_w, t_h]_\two, [t_x, t_y, t_z]_\mathbf{P}, [t_w, t_h, t_l, t_\theta]_\three$. 
Each output uses a $1\times1$ kernel and are collectively denoted as $\mathbf{O}_\text{global}$ and $\mathbf{O}_\text{local}$.
To leverage the depth-aware and spatial-invariant strengths, we fuse each output using a learned attention $\alpha$ (after sigmoid) applied for $i=1\dots 12$ as follows:
\begin{equation}
\eqnvspace
\mathbf{O}^i = \mathbf{O}^i_\text{global} \cdot \alpha_i + \mathbf{O}^i_\text{local} \cdot (1 - \alpha_i).
\eqnvspace
\end{equation}
%The $c$ classification output generates  $(n_a \times n_c)$ features where $n_a$ denotes number of anchors and $n_c$ number of classes. 
%All other remaining outputs generate $n_a$ feature maps. 
%where spatial-invariant convolution and depth-aware convolution are respectively utilized for the $\mathbf{O}_\text{global}$ and $\mathbf{O}_\text{local}$ outputs, as illustrated in detail within Fig.~\ref{fig:overview}.

\SubSection{Post 3D$\rightarrow$2D Optimization}
We optimize the orientation parameter $\theta$ in a simple but effective post-processing algorithm (as detailed in Alg.~\ref{hillclimb}).
The proposed optimization algorithm takes as input both the 2D and 3D box estimations $b'_\two$, $[x, y, z]'_\mathbf{P}$, and $[w, h, l, \theta]'_\three$, as well as a step size $\sigma$, termination $\beta$, and decay $\gamma$ parameters. 
The algorithm then iteratively steps through $\theta$ and compares the projected 3D boxes with $b'_\two$ using a $L_1$ loss. 
The 3D$\rightarrow$2D box-project function is defined as follows:
\vspace{-.0mm}
\begin{equation}\vspace{1.5mm}
\begin{aligned}
\Upsilon_0 &= 
\setlength\arraycolsep{3.1pt}
\begin{bmatrix}
    -l &  l &  l & l &  l & -l & -l & -1  \\
    -h & -h &  h & h & -h & -h &  h &  h \\
    -w & -w & -w & w &  w &  w &  w & -w \\
  \end{bmatrix}'_\three/~2,\\
\Upsilon_\three &= \begin{bmatrix}
    \cos\theta & 0 & \sin\theta\\
    0 & 1 & 0 \\
    -\sin\theta & 0 & \cos\theta \\
    0 & 0 & 0 \\
  \end{bmatrix} 
  \Upsilon_0 + \mathbf{P}\inv \begin{bmatrix}
    x\cdot z \\
    y\cdot z \\
    z \\
    1
  \end{bmatrix}'_\mathbf{P}, \\ 
  \Upsilon_\mathbf{P} &= \mathbf{P} \cdot \Upsilon_\three ,~~~~~~~~~~~~~~
  \Upsilon_\two = \Upsilon_\mathbf{P}./ \Upsilon_\mathbf{P}[\phi_z], \\
 x_\text{min} &= \min(\Upsilon_\two[\phi_x]),~~~~y_\text{min} = \min(\Upsilon_\two[\phi_y]), \\
 x_\text{max} &= \max(\Upsilon_\two[\phi_x]),~~~ y_\text{max} = \max(\Upsilon_\two[\phi_y]).
\end{aligned}
\end{equation}
where $\mathbf{P}\inv$ is the inverse projection after padding $[0, 0, 0, 1]$, and $\phi$ denotes an index for axis $[x, y, z]$.
We then use the projected box parameterized by $\rho = [x_\text{min}, y_\text{min}, x_\text{max}, y_\text{max}]$ and the source $b'_\two$ to compute a $L_1$ loss, which acts as the driving heuristic. % to optimize $\theta$. % to optimize using a basic gradient hill-climbing method. 
When there is no improvement to the loss using $\theta \pm \sigma$, we decay the step by $\gamma$ and repeat while $\sigma \geq \beta$. %, as detailed in Fig.~\ref{hillclimb}.
%optimizing $\theta$ through an IoU-based loss computed between the $b'_\two$ and a projected $b'_\three$. 
%Specifically, the we convert the 3D box into 2D bounding box by projecting the $8$ 3D corners into the image space using the camera projection matrix $\mathbf{P}$, then taking the min and max in both the 2D $X$ and $Y$ axis to define the 2D box counterpart. 

\SubSection{Implementation Details}
We implement our framework using PyTorch~\cite{paszke2017automatic} and release the code at \small\url{http://cvlab.cse.msu.edu/project-m3d-rpn.html}\normalsize. 
To prevent local features from overfitting on a subset of the image regions, we initialize the local path with pretrained global weights.
%Hence, we first train with the local and global paths sharing kernels, then transfer the local path into depth-aware kernels and re-train. 
%Then we re-train after initializing the local path using the global kernels. 
In this case, each stage is trained for $50k$ iterations. 
We expect higher degrees of data augmentation or an iterative binning schedule, e.g., $b=2^i$ from $i=0\dots \log_2(b_\text{final})$, could enable more ease of training at the cost of more complex hyperparameters.  

We use a learning rate of $0.004$ with a poly decay rate using power $0.9$, a batch size of $2$, and weight decay of $0.9$.
We set $\lambda_1 = \lambda_2 = 1$. 
All images are scaled to a height of $512$ pixels. 
As such, we use $b = 32$ bins for all depth-aware convolution layers.
We use $12$ anchor scales ranging from $30$ to $400$ pixels following the power function of $30\cdot 1.265 ^i$ for $i=0 \dots 11$ and aspect ratios of $[0.5, 1.0, 1.5]$ to define a total of $36$ anchors for multi-class detection. 
The 3D anchor priors are learned using these templates with the training dataset as detailed in Sec.~\ref{sec:formulation}.
We apply NMS on the box outputs in the 2D space using a IoU criteria of $0.4$ and filter boxes with scores $< 0.75$. 
The 3D~$\to$~2D optimization uses settings of $\sigma=0.3 \pi$, $\beta=0.01$, and $\gamma=0.5$.
Lastly, we perform random mirroring and online hard-negative mining by sampling the top $20\%$ high loss boxes in each minibatch. 

We note that M3D-RPN relies on 3D box annotations and a known projection matrix $\mathbf{P}$ per sequence.
For extension to a dataset without these known, it may be necessary to predict the camera intrinsics and utilize weak supervision leveraging 3D-2D projection geometry as loss constraints. 

%% file: sec_4.tex
\section{Experiments} 

\begin{figure*}[t]
\vspace{-2mm}
\begin{center}
   \includegraphics[width=0.99\linewidth]{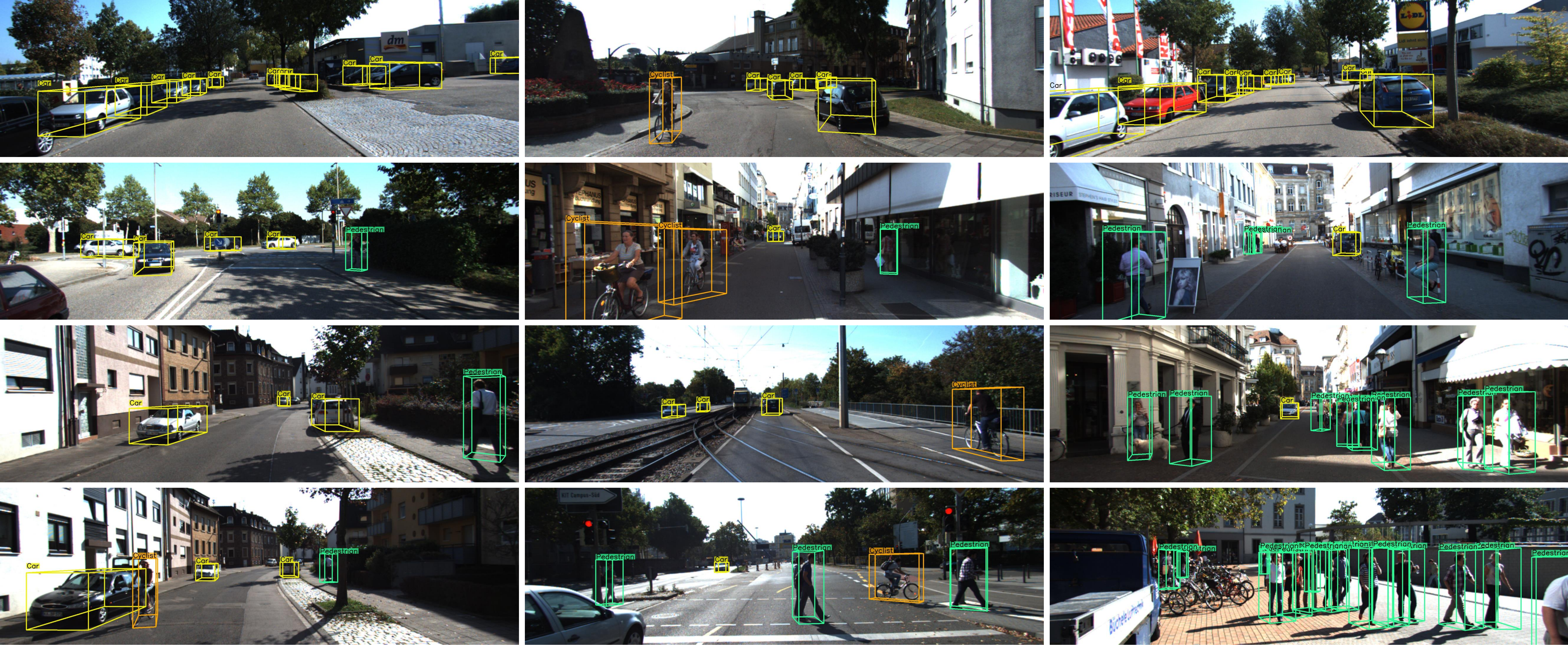}
\vspace{-0mm}
      \caption{
\textbf{Qualitative Examples}. We visualize qualitative examples of our method for multi-class 3D object detection. 
We use yellow to denote cars, green for pedestrians, and orange for cyclists. All illustrated images are from the val1~\cite{chen20153d} split and {not} used for training.
}
\label{fig:qual}
\end{center}\vspace{-6mm}
\end{figure*}

We evaluate our proposed framework on the challenging KITTI~\cite{Geiger2012CVPR} dataset under two core 3D localization tasks: Bird's Eye View (BEV) and 3D Object Detection. 
We comprehensively compare our method on the official test dataset as well as two validation splits~\cite{chen20153d, xiang2017subcategory}, and perform analysis of the critical components which comprise our framework.
We further visualize qualitative examples of M3D-RPN on multi-class 3D object detection in diverse scenes (Fig.~\ref{fig:qual}).

\SubSection{KITTI}
The KITTI~\cite{Geiger2012CVPR} dataset provides many widely used benchmarks for vision problems related to self-driving cars.
Among them, the Bird's Eye View (BEV) and 3D Object Detection tasks are most relevant to evaluate 3D localization performance.
%Among them, the BEV and 3D Object Detection standout as the foremost critical tasks best suited to evaluate the performance of 3D recognition and localization. 
The official dataset consists of $7{,}481$ training images and $7{,}518$ testing images with 2D and 3D annotations for car, pedestrian, and cyclist.  
For each task we report the Average Precision (AP) under $3$ difficultly settings: easy, moderate and hard as detailed in~\cite{Geiger2012CVPR}.
Methods are further evaluated using different IoU criteria per class. 
We emphasize our results on the official settings of IoU $\geq0.7$ for cars and IoU $\geq 0.5$ for pedestrians and cyclists.

We conduct experiments on three common data splits including val1~\cite{chen20153d}, val2~\cite{xiang2017subcategory}, and the official test split~\cite{Geiger2012CVPR}. 
Each split contains data from non-overlapping sequences such that no data from an evaluated frame, or its neighbors, have been used for training. 
%For experiments conducted on the official test dataset we use the corresponding full training dataset for supervision.
%For validation experiments we denote the train-val split proposed in~\cite{chen20153d} as {val$1$} and the split of~\cite{xiang2017subcategory} as {val$2$}.
%Each split separates the official training set approximately in half while avoiding overlap from the raw KITTI sequences. 
We focus our comparison to SOTA prior work which use image-only input. %, including stereo and monocular. 
We primarily compare our methods using the car class, as has been the focus of prior work~\cite{chen2016monocular, chen20153d, mousavian20173d, xu2018multi}. 
However, we emphasize that our models are trained as a shared multi-class detection system and therefore also report the multi-class capability for monocular 3D detection, as detailed in Tab.~\ref{tab:MULTI}.
%However, we further introduce multi-class results in Tab.~\cite{?}.

\begin{table} [t]
    \begin{center}
\small
\setlength\tabcolsep{4.25pt}
\begin{tabular}{c | c | c }
\hline
 & \multicolumn{1}{c|}{AP$_\text{BEV}$~~\small{[val1~/~val2~/~test]}} & \multicolumn{1}{c}{AP$_\text{3D}$~~\small{[val1~/~val2~/~test]}}  \\ 
\hline
Car & $21.18$~/~$21.15$~/~$18.36$ & $17.06$~/~$16.48$~/~$15.70$ \\
Pedestrian  & $11.60$~/~$11.44$~/~$11.35$ & $11.28$~/~$11.30$~/~$10.54$ \\
Cyclist  & $10.13$~/~$~9.09$~/~$~1.29$ & $10.01$~/~$~9.09$~/~$~1.03$ \\
\hline
\end{tabular}
\vspace{1mm}
\caption{\textbf{Multi-class 3D Localization}. The performance of our method when applied as a multi-class 3D detection system using a single shared model. 
We evaluate using the {mod} setting on KITTI. 
}\label{tab:MULTI}
\end{center}\vspace{-4mm}
\end{table}

\begin{table} [t]
    \begin{center}
\small
\setlength\tabcolsep{3.1pt}
\begin{tabular}{c | c c c }
\hline
 & \multicolumn{3}{c}{2D Detection [val1 / test]}  \\ 
& Easy & Mod & Hard  \\ 
\hline
Mono3D~\cite{chen2016monocular} & $93.89$ / $92.33$ & $88.67$ / $88.66$ & $79.68$ / $78.96$ \\
3DOP~\cite{chen20153d} & $93.08$ / $93.04$ & $88.07$ / $88.64$ & $79.39$ / $79.10$ \\
Deep3DBox~\cite{mousavian20173d}  & ~~~~-~~~~~/ $92.98$ & ~~~~-~~~~~/ $89.04$ & ~~~~-~~~~~/ $77.17$ \\
Multi-Fusion~\cite{xu2018multi}  & ~~~~-~~~~~/ $90.43$ & ~~~~-~~~~~/ $87.33$ & ~~~~-~~~~~/ $76.78$ \\
\hline \hline
M3D-RPN & $90.24$ / $84.34$ & $83.67$ / $83.78$ & $67.69$ / $67.85$ \\
\hline 
\end{tabular}
\vspace{1mm}
\caption{
\textbf{2D Detection}. The performance of our method evaluated on 2D detection using the car class on val1 and test datasets.  
}\label{tab:2D}
\end{center}\vspace{-4mm}
\end{table}

\Paragraph{Bird's Eye View:}
The Bird's Eye View task aims to perform object detection from the overhead viewpoint of the ground plane.
Hence, all 3D boxes are first projected onto the ground plane then top-down 2D detection is applied. 
%The task is effectively equivalent to 2D detection conducted using 3D ground truths boxes projected onto the $XZ$ ground-plane.
%Thus, the effect of the estimating $y_\text{3D}$ and $h_\text{3D}$ is essentially removed.
%However it n autonomous driving scenarios the Bird's Eye View is arguably the most important viewpoint for planning~\cite{?}. 
We evaluate M3D-RPN on each split as detailed in Tab.~\ref{tab:BEV}. % as shown in detail in Tab.~\ref{?}.

M3D-RPN achieves a notable improvement over SOTA image-only detectors across all data splits and protocol settings. %using test, val$1$, and val$2$. 
For instance, under criteria of IoU $\geq 0.7$ with val1, our method achieves $21.18\%$ ($\uparrow 7.55\%$) on moderate, and $17.90\%$ ($\uparrow 6.30\%$) on hard. % which is proposed setting in KITTI~\cite{Geiger2012CVPR}. % which is among the most difficult of 3D localization test settings evaluated for the Bird's Eye View task.
%We also observe similar performance on the other splits. 
We further emphasize our performance on test which achieves $18.36\%$ ($\uparrow 8.74\%$) and $16.24\%$ ($\uparrow 8.02\%$) respectively on moderate and hard settings with IoU $\geq 0.7$, which is the most challenging setting.

\Paragraph{3D Object Detection:}
The 3D object detection task aims to perform object detection {directly} in the camera coordinate system. 
Therefore, an additional dimension is introduced to all IoU computations, which substantially increases the localization difficulty compared to BEV task. 
We evaluate our method on 3D detection with each split under all commonly studied protocols as described in Tab.~\ref{tab:3D}.
Our method achieves a significant gain over state-of-the-art image-only methods throughout each protocol and split. 

We emphasize that the current most {difficult} challenge to evalaute 3D localization is the 3D object detection task. 
Similarly, the moderate and hard settings with IoU $\geq 0.7$ are the most difficult protocols to evaluate with.
Using these settings with val1, our method notably achieves $17.06\%$ ($\uparrow 11.37\%$) and $15.21$ ($\uparrow 9.82\%$) respectively. 
We further observe similar gains on the other splits.
For instance, when evaluated using the testing dataset, we achieve $15.70\%$ ($\uparrow 10.52$) and $13.32\%$ ($\uparrow 8.64$) on the moderate and hard settings {despite} being trained as a \textit{shared} multi-class model and compared to single model methods~\cite{chen2016monocular, chen20153d, mousavian20173d, xu2018multi}. 
When evaluated with less strict criteria such as IoU $\geq 0.5$, our method demonstrates smaller but reasonable margins ($\Sim 3-6\%$), implying that M3D-RPN has similar recall to prior art but significantly higher {precision} overall. 
%Interestingly, most methods perform comparably under the IoU$\geq 0.5$ settings. 
%Thus, our proposed framework is more robust as the localization criteria becomes more challenging. 
%As shown in Fig.~\ref{?}, our proposed framework also achieves a significant gain over the state-of-the-art detectors from $X \to X$. 
%Specifically, our framework improves performance by a factor of $3\times$. 

\Paragraph{Multi-Class 3D Detection:}
To demonstrate generalization beyond a single class, we evaluate our proposed  3D detection framework on the car, pedestrian and cyclist classes. 
We conduct experiments on both the Bird's Eye View and 3D Detection tasks using the KITTI test dataset, as detailed in Tab.~\ref{tab:MULTI}. 
%It is worth noting that multi-class single models are uncommon in urban detection in general as most state-of-the-art methods~\cite{?, ?, ?, ?} train separate detectors on a per class basis. 
%Further, no prior work report performance on multi-class for either 3D task in question.
%We report our performance on each task as summarized in Tab.~\ref{?}. 
Although there are not monocular 3D detection methods to compare with for multi-class, it is noteworthy that the performance on {pedestrian} outperforms prior work performance on \textit{car}, which usually has the opposite relationship, thereby suggesting a reasonable performance. 
However, M3D-RPN is noticeably less stable for cyclists, suggesting a need for advanced sampling or data augmentation to overcome the data bias towards {car} and {pedestrian}.

\Paragraph{2D Detection:}
We evaluate our performance on 2D car detection (detailed in Tab.~\ref{tab:2D}). 
We note that M3D-RPN performs less compared to other 3D detection systems applied to the 2D task.
However, we emphasize that prior work~\cite{chen2016monocular, chen20153d, mousavian20173d, xu2018multi} use external networks, data sources, and include multiple stages (e.g., Fast~\cite{girshick2015fast}, Faster R-CNN~\cite{ren2015faster}). 
In contrast, M3D-RPN performs all tasks simultaneously using only a single-shot 3D proposal network.
Hence, the focus of our work is primarily to improve 3D detection proposals with an emphasis on the quality of 3D localization.
%We observe that the during training the 3D loss typically dominates 
Although M3D-RPN does not compete directly with SOTA methods for 2D detection, its performance is suitable to facilitate the tasks in focus such as BEV and 3D detection.
%The relative effect of our method with respect to 2D detection is lightly revealed as a side-effect of our ablations described in Sec.~\ref{sec:ablations} and Tab.~\ref{tab:ablations}.

\SubSection{Ablations}\label{sec:ablations}

\begin{table} [t]
    \begin{center}
    \small
\setlength\tabcolsep{5.pt}
\begin{tabular}{c c | c c c | c }
\hline
$b$  & Post-Optim & AP$_\text{2D}$ & AP$_\three$ & AP$_\text{BEV}$ &  RT (ms) \\ 
\hline
 &  & $82.16$& $10.99$& $12.99$ & $118$\\
 & $\checkmark$ & $82.16$& $15.08$& $17.47$ & $128$  \\
$1$ & $\checkmark$ & $82.88$& $12.87$& $17.91$ &  $133$ \\
$4$ & $\checkmark$ & $84.15$& $14.46$& $19.14$ &  $134$ \\
$8$ & $\checkmark$ & $83.86$& $16.04$& $20.99$ & $143$ \\
$16$ & $\checkmark$ & $83.02$& $15.97$ & $18.48$ & $153$  \\
\hline \hline
$32$ & $\checkmark$ & $83.67$& $17.06$& $21.18$ & $161$ \\
\hline
\end{tabular}
\vspace{1mm}
\caption{\textbf{Ablations}. We ablate the effects of $b$ for depth-aware convolution and the post-optimization 3D$\to$2D algorithm with respect to performance on moderate setting of cars and runtime (RT). 
}\label{tab:ablations}
\end{center}\vspace{-4mm}
\end{table}

\begin{table} [t]
    \begin{center}
    \small
\setlength\tabcolsep{3.pt}
\begin{tabular}{c c c c c c c c c c c c c c }
\hline
$c$ & $x_\two$ & $y_\two$ & $w_\two$ & $h_\two$  & $x_\mathbf{P}$ & $y_\mathbf{P}$ & $z_\mathbf{P}$ & $w_\three$ & $h_\three$ & $l_\three$ & $\theta_\three$ &  \\
$33$ & $48$ & $47$ & $45$ & $45$ & $44$ & $45$ & $44$ & $42$ & $38$ & $43$ & $38$ & $\%$ \\
\hline
\end{tabular}
\vspace{1mm}
\caption{\textbf{Local and Global $\alpha$ weights}. We detail the $\alpha$ weights learned to individually fuse each global and local output.
Lower implies higher weight towards the local depth-aware convolution. %whereas higher implies preference for the global outputs. 
}\label{tab:breakdown}
\end{center} \vspace{-1mm}
\end{table}

For all ablations and experimental analysis we use the KITTI val1 dataset split and evaluate utilizing the car class. 
Further, we use the moderate setting of each task which includes 2D detection, 3D detection, and BEV (Tab.~\ref{tab:ablations}). 
%We further provide discussion and analysis on efficiency. 
%We evaluate the effect of depth-aware convolutoin and the post-optimization algorithm via ablation experiments as summarized in omponent of our proposed method as it contributes to the overall performance.
%To do so, we cumulatively remove a single component at at time as is summarized within Tab.~\ref{?}.
%All ablations are conducted on the KITTI val$1$~\cite{?} split following the proposed settings outlined in Sec.~\ref{?}, unless otherwise stated. 

\Paragraph{Depth-aware Convolution:}
We propose depth-aware convolution as a method to improve the spatial-awareness of high-level features. 
%In doing so, the network is able to learn specific kernels and features dependent on the discretized level of depth. 
To better understand the effect of depth-aware convolution, we ablate it from the perspective of the hyperparameter $b$ which denotes the number of discrete bins. 
Since our framework uses an image scale of $512$ pixels with network stride of $16$, the output feature map can naturally be separated into $\frac{512}{16} = 32$ bins.
We therefore ablate using bins of $[4, 8, 16, 32]$ as described in Tab.~\ref{tab:ablations}. 

We additionally ablate the special case of $b=1$, which is the equivalent to utilizing two global streams. 
We observe that both $b=1$ and $b=4$ result in generally worse performance than the baseline without local features, suggesting that arbitrarily adding deeper layers is not inherently helpful for 3D localization. 
However, we observe consistent improvements when $b=32$ is used, achieving a large gain of $3.71\%$ in AP$_\text{BEV}$, $1.98\%$ in AP$_\three$, and $1.51\%$ in AP$_\two$.

We breakdown the learned $\alpha$ weights after sigmoid which are used to fuse the global and local outputs (Tab.~\ref{tab:breakdown}). 
Lower values favor local branch and vice-versa for global.  
Interestingly, the classification $c$ output learns the highest bias toward local features, suggesting that semantic features in urban scenes have a moderate reliance on depth position. 
%We observe that as more bins are used the performance generally increases on most of the evaluated tasks.
%In the extreme case between no depth-aware convolution is used the  and full depth-aware convolution ($b=32$), the network ears a $X\%$ mAP gain in performance. 
%This gain emphasizes the usefulness of depth-aware convolution.  
%We note that the overall network memory footprint does increase as $b$ grows larger, depicted in Tab.~\ref{?}.
%However, we feel the gain of performance warrants the slight ($\Sim X\%$) increase in GPU memory. 
%Furthermore, future work of non-uniform implementation of depth-aware convolution, where bins each use a custom split may further help memory efficiency. 

\Paragraph{Post 3D$\rightarrow$2D Optimization:}
The post-optimization algorithm encourages consistency between 3D boxes projected into the image space and the predicted 2D boxes. 
%The algorithm uses a straightforward $L_1$ corner loss with a greedy hill-climbing method (as detailed in Alg.~\ref{hillclimb}) to improve the orientation $\theta$ estimation. 
We ablate the effectiveness of this optimization as detailed in Tab.~\ref{tab:ablations}. 
We observe that the post-optimization has a significant impact on both BEV and 3D detection performance. 
Specifically, we observe performance gains of $4.48\%$ in AP$_\text{BEV}$ and $4.09\%$ in AP$_\three$. 
We additionally observe that the algorithm converges in approximately $8$ iterations on average and adds minor $13$ ms overhead (per image) to the runtime. 

\Paragraph{Efficiency:}
We emphasize that our approach uses only a single network for inference and hence involves overall more direct 3D predictions than the use of multiple networks and stages (RPN with R-CNN) used in prior works ~\cite{chen2016monocular, chen20153d, mousavian20173d, xu2018multi}. 
We note that direct efficiency comparison is difficult due to a lack of reporting in prior work.
However, we comprehensively report the efficiency of M3D-RPN for each ablation experiment, where $b$ and post-optimization are the critical factors, as detailed in Tab.~\ref{tab:ablations}. 
The runtime efficiency is computed using NVIDIA $1080$ti GPU averaged across the KITTI val1 dataset. 
We note that depth-aware convolution incurs $2-20\%$ overhead cost for $b=1 \dots 32$, caused by unfolding and reshaping in PyTorch~\cite{paszke2017automatic}. 
%To better understand the effects of this algorithm, we examine the IoU consistency, rotation estimation error, and performance with and without the algorithm. 
%We conduct this study using the val$1$ split across all ground truths matched with a prediction. 
%Firstly, we observe that the 3D projected IoU consistency with its 2D counterpart increases from a mean of $X\%$ to $X\%$.
%The greedy algorithm takes approximately $\Sim X$ iterations to converge per box and improves the estimated rotation error by $X$rad on average. 
%Most importantly, we observe when the post optimization is activated for the core M3D-RPN method, the performance increases from $X \to X$ mAP on the moderate setting of with IoU $\geq 0.7$, reaffirming its usefulness in practice.  
%Lastly, we visualize the per iteration changes caused by the algorithm for the most extreme cases (top $X\%$ boxes with respect to $\Delta \theta$), illustrated in Fig.~\ref{?}.

%% file: sec_5.tex
\section{Conclusion}

In this work, we present a reformulation of monocular image-only 3D object detection using a \textit{single-shot} 3D RPN, in contrast to prior work which are comprised of external networks, data sources, and involve multiple stages.
M3D-RPN is uniquely designed with shared 2D and 3D anchors which leverage strong priors closely linked to the correlation between 2D scale and 3D depth. 
%Our proposed framework is uses only a single shared multi-class network in comparison to prior work comprised of multiple external networks. 
To help improve 3D parameter estimation, we further propose depth-aware convolution layers which enable the network to develop spatially-aware features. 
Collectively, we are able to significantly improve the performance on the challenging KITTI dataset on both the Birds Eye View and 3D object detection tasks for the car, pedestrian, and cyclist classes. 

\Paragraph{Acknowledgment:}
Research was partially sponsored by the Army Research Office under Grant Number W911NF-18-1-0330. The views and conclusions contained in this
document are those of the authors and should not be interpreted as representing the official policies, either expressed or implied, of the Army Research Office or the U.S. Government. The U.S. Government is authorized to reproduce and distribute reprints for Government purposes notwithstanding any copyright notation herein.